\documentclass{article}

\usepackage{microtype}
\usepackage{graphicx}
\usepackage{subfigure}
\usepackage{booktabs} 

\usepackage{amsmath, array, mleftright}
\usepackage{amssymb}
\usepackage{bbold}
\usepackage{xparse}
\usepackage[english]{babel}
\usepackage{amsthm}
\usepackage{color}
\usepackage{lipsum}
\usepackage{forest}

\usepackage{hyperref}

\usepackage[accepted]{icml2018}

\icmltitlerunning{DCRNN}

\NewDocumentCommand{\evalat}{sO{\big}mm}{%
  \IfBooleanTF{#1}
   {\mleft. #3 \mright|_{#4}}
   {#3#2|_{#4}}%
}

\newcommand{\norm}[1]{\left\lVert#1\right\rVert}

\usepackage{xspace}
\newcommand{\rnn}{{DCRNN}\xspace}

\begin{document}

\twocolumn[
\icmltitle{A Dynamically Controlled Recurrent Neural Network\\  for Modeling Dynamical Systems}




\icmlsetsymbol{equal}{*}

\begin{icmlauthorlist}
\icmlauthor{Yiwei Fu$^*$}{me}
\icmlauthor{Samer Saab Jr$^*$}{ee}
\icmlauthor{Asok Ray}{me}
\icmlauthor{Michael Hauser}{chop}
\end{icmlauthorlist}

\icmlaffiliation{me}{The Pennsylvania State University, Department of Mechanical Engineering}
\icmlaffiliation{ee}{The Pennsylvania State University, Department of Electrical Engineering}
\icmlaffiliation{chop}{Children's Hospital of Philadelphia}

\icmlcorrespondingauthor{Michael Hauser}{mikebenh@gmail.com}

\icmlkeywords{Recurrent Neural Network, Dynamical System, Ordinary Differential Equation, Lyapunov Stability, State-Space Model, Lorenz System}

\vskip 0.3in
]



\printAffiliationsAndNotice{\icmlEqualContribution} 

\begin{abstract}

This work proposes a novel neural network architecture, called the Dynamically Controlled Recurrent Neural Network (\rnn), specifically designed to model dynamical systems that are governed by ordinary differential equations (ODEs). The current state vectors of these types of dynamical systems only depend on their state-space models, along with the respective inputs and initial conditions. Long Short-Term Memory (LSTM) networks, which have proven to be very effective for memory-based tasks, may fail to model physical processes as they tend to memorize, rather than learn how to capture the information on the underlying dynamics. The proposed \rnn includes learnable skip-connections across previously hidden states, and introduces a regularization term in the loss function by relying on Lyapunov stability theory. The regularizer enables the placement of eigenvalues of the transfer function induced by the \rnn to desired values, thereby acting as an internal controller for the hidden state trajectory. The results show that, for forecasting a chaotic dynamical system, the \rnn outperforms the LSTM in $100$ out of $100$ randomized experiments by reducing the mean squared error of the LSTM's forecasting by $80.0\% \pm 3.0\%$.

\end{abstract}

\section{Introduction}

Recurrent neural networks have set the state-of-the-art on an enormous array of diverse time series tasks \cite{lipton2015critical}. They are designed for processing time series by sharing weights through time. It is remarkable that they are so successful at processing general time series, even though time series can be generated by systems of an extremely diverse nature. For example, the mechanisms involved behind a time series generated by a mass-spring-damper system are incomparably different from linguistic time series produced by a generative grammar \cite{chomsky2002syntactic}. This work will distinguish time series as those having been produced by a dynamical system from those that have been produced by a system with an internal memory. Our work designs a novel and easily interpretable recurrent network, based on a standard technique in control theory, namely the Lyapunov linearization method \cite{lyapunov1949probldme}.

In cases where the system has an internal memory, there exists many difficulties in training recurrent networks, such as learning long-term dependencies between the inputs \cite{bengio1994learning, pascanu2013difficulty}. This is a quintessential difficulty in automatic speech recognition \cite{ravanelli2018light, chiu2018state}.

Several variants of the recurrent network have been proposed to alleviate the aforementioned problems, such as the Long Short-Term Memory network, or LSTM \cite{hochreiter1997long, gers1999learning}; which have gained wide popularity and have been thoroughly studied \cite{jozefowicz2015empirical, greff2017lstm}. Various architectures have been proposed as extensions of LSTMs, such as the forget-gate only architecture introduced called the JANET \cite{van2018unreasonable}. These efforts are heavily dependent on memory and contain a memory-cell, which exists in order to retain information from the past and attenuates the effects of vanishing/exploding gradients. 

In the case that the input was generated by a dynamical system without memory, such as systems modeled by differential equations, LSTM-like networks will still try to use their memory cell to remember the trajectory of the inputs, instead of learning the underlying equations governing the trajectory of the system. The fundamental problem of modeling dynamical systems with an LSTM is that differential equations do not have a long-term memory, they are only dependent on the current state, input and the equations governing the state evolution. Therefore, LSTM models often try to find the long-term memory dependencies when they do not actually exist.

Lyapunov's linearization method is a mathematical technique developed to study the stability of nonlinear systems \cite{kalman1960control}. This method allows us to determine the stability of nonlinear systems in the neighborhood of equilibrium points on the basis of the linearized system. In the context of recurrent networks, we treat the hidden states of the recurrent network as the system's state variables, which is important as this allows us to control their trajectories as they propagate forward in time. If the hidden states are unstable, this prevents the network from learning as it saturates its corresponding activation functions and thus leads to vanishing gradients for backpropagation.

With these motivations we develop the Dynamically Controlled Recurrent Neural Network, or \rnn, which is designed in a way such that Lyapunov's linearization methods are used to control the trajectories of the hidden states. The key contributions of this paper are:
\vspace{-0.2cm}
\begin{itemize}
    \item Introduce hidden state connections, with learnable coefficients, over a number of previous time steps.
    \item Construct a nonlinear state-space representation of the network; which is then linearized using Lyapunov's linearization method. The state variables considered are the hidden-states of the recurrent network.
    \item Control the placement of the eigenvalues of the linearized system by introducing a regularization term to the overall cost function, compelling the eigenvalues to converge to their desired positions during training.
\end{itemize}
\vspace{-0.2cm}
The regularizer acts as a controller to the system in the context of control theory, a concept that we have not seen implemented before, to the best of our knowledge. Neural network architectures exist that introduce gating functions that the designers call a controller cell, for example \cite{tay2018controlled}, with the goal that the neural network \emph{learn} the controller during training. In our design, we \emph{impose} a controller in the \rnn that is specifically designed to stabilize the hidden state trajectory through eigenvalue placement, making it consistent with the term controller adapted in control theory.

The \rnn is a generalization of the vanilla RNN and can also be parameterized to become the constant error carousel structure in LSTM, thus providing greater model flexibility and better representation power.


\section{Related Work}
Our neural architecture, the \rnn, is meant for (i) modeling dynamical systems; it focuses on (ii) controlling hidden states within RNNs, and (iii) uses skip-connections in order to model differential equations of various orders. This section briefly highlights previous work that concerns these three areas.
\vspace{-0.2cm}
\paragraph{Modeling Dynamical Systems} Various efforts have been introduced to deal with modeling dynamical systems using RNNs. 
A generative approach was recently proposed to model time-sereies data by introducing an additional step to an RNN autoencoder~\cite{NIPS2018_7892}. This approach takes as input time-series data, along with the initial state of the data, and uses an external ordinary differential equations (ODE) solver to produce latent state variables that describes a unique latent trajectory of the input. Using this uniquely defined trajectory, predictions can be made arbitrarily far into the future or past. Neitz et. al~\citeyearpar{neitz2018adaptive} introduced the Adaptive Skip Intervals (ASI), which classifies tasks where the labels are generated by dynamical systems. This method adjusts the time-frame at which predictions are made, as the goal is to capture the dynamics of the observed input in terms of causal mechanisms; similar to the causal inference approach. This approach is built to have varying prediction intervals as it searches for the easiest time frames in which the input can be correctly classified, whereas the \rnn is designed to control trends in the hidden space needed to make an accurate prediction at every time-step.
\vspace{-0.2cm}
\paragraph{Containing Hidden States} Similar to our efforts in controlling the hidden-states around acceptable values is recurrent batch normalization \cite{cooijmans2016recurrent}, where hidden states are scaled and normalized according to the layer mean and variance. Through this approach, the hidden states (and input) are essentially zero-meaned and thus are maintained at an acceptable value such that they do not diverge allowing the RNN to learn. The network however is sensitive to initialization of the batch-normalization (BN) network parameters, and vulnerable to overfitting in certain applications, as indicated in the paper. BN was also applied to RNNs in \cite{laurent2016batch, liao2016bridging, amodei2016deep}, but was found not to improve results when applied to hidden states.
\vspace{-0.2cm}
\paragraph{Skip-Connections in RNNs} Chang et al.~\citeyearpar{chang2017dilated} introduced the DilatedRNN, where the number of time-steps skipped, or dilation, grows exponentially as the number of layers is increased. The DilatedRNN achieves state-of-the-art results even when using vanilla RNN cells. Another approach was to design a hierarchical RNN~\cite{el1996hierarchical}, which introduces several levels of state variables operating at different time-scales through discrete time-step delays. As the number of different time-scales considered in the network was increased, the test errors decreased for long-sequence data. However, both cases use an identity mapping for the skip connections, which as discussed above may very likely lead the hidden states to blow-up in value and saturate in their corresponding activation functions. 

As mentioned earlier, the term controller used in this paper is consistent with control theory and is different from the term \textit{controller} in \cite{tay2018controlled} which represents a cell that learns gating functions in order to influence their target cell.

\section{Proposed Approach}\label{sec:propose}

To improve and control the flow of information through an RNN, we introduce the \rnn architecture which contains a parameter $k \in \mathbb{Z}^{+}$ that refers to the number of \textit{weighted} skip connections across time-steps, in the following manner: 
\begin{equation}
    h_{t} = \sum_{i=1}^{k}\big\{\alpha_{i}h_{t-i}\big\} + \varphi_h(h_{t-1},x_{t})
    \label{rnn_architecture}
\end{equation}
where $\alpha_i \in \mathbb{R}^{n \times n}, \forall i = 1,\hdots, k$ is a diagonal matrix with diagonal entries $\alpha_{i}^{(j)}, \forall j = 1, \hdots, n$, and $\varphi_{h}( \cdot )$ is the activation function. The matrices $\alpha_{i}$ are referred to as the skip-coefficients, $h_{t} \in \mathbb{R}^{n}$ are the hidden states commonly used in RNNs, and $x_{t} \in \mathbb{R}^{n}$ is the input to the RNN. Notice that if we set $k=0$, the system reduces to a vanilla RNN. Whereas for $k=1$ and $\alpha_{1}=\mathbb{1}$, the system reduces to the constant error carousel structure in the original LSTM paper~\cite{hochreiter1997long}. It should also be noted that the forget gate was added to the LSTM later~\cite{gers1999learning}, effective by making $\alpha_1=f_t$, where $f_t$ is the forget gate value. 

Furthermore, we let the skip-coefficients be learnable parameters in order to allow the network to vary the coefficients such that the the states of the RNN are controlled. To achieve such control, we introduce a regularizer that is added to the loss-function which updates the skip-coefficients such that the states of the RNN never diverge. In order to do so, we create a linearized state-space representation of the network and build the regularizer to ensure that the absolute value of the eigenvalues of the linearized state-space model are inside the unit circle. 


\subsection{State-Space Representation of \rnn Architecture}
With the architecture described above, it becomes possible to define state-vectors as all previous connected hidden states, in the following manner $q_{t}^{j} := h_{t-j}, \forall j = 1, \hdots, k$. Thus, we can create a state-space formulation describing the hidden states' behavior as:

\begin{equation}
\resizebox{0.45\textwidth}{!}{$
\begin{aligned}
\begin{bmatrix}
q_{t+1}^{1} \\ q_{t+1}^{2} \\ \vdots \\ q_{t+1}^{k-1} \\ q_{t+1}^{k}
\end{bmatrix}
&=
\begin{bmatrix}
\alpha_{1} & \alpha_{2} & \hdots &  \alpha_{k-1} & \alpha_{k} \\
\mathbb{1} & \mathbb{0} & \hdots & \mathbb{0} & \mathbb{0} \\
\mathbb{0} & \mathbb{1} & \hdots & \mathbb{0} & \mathbb{0} \\
\vdots & \ddots  & \ddots & \vdots & \vdots \\
\mathbb{0} & \mathbb{0} & \hdots &  \mathbb{1} & \mathbb{0} \\
\end{bmatrix}
\begin{bmatrix}
q_{t}^{1} \\ q_{t}^{2} \\ \vdots \\ q_{t}^{k-1} \\ q_{t}^{k}
\end{bmatrix}
+
\begin{bmatrix}
\mathbb{1} \\ \mathbb{0} \\ \vdots \\ \mathbb{0} \\ \mathbb{0}
\end{bmatrix}
\varphi_h(q_{t}^{1},x_{t})
\\
&\triangleq
\begin{bmatrix}
f^{1}(q^{1}_{t}, q^{2}_{t}, \hdots, q^{k}_{t}, x_{t}) \\ f^{2}(q^{2}_{t}, x_{t}) \\ \vdots \\ f^{k-1}(q^{k-1}_{t}, x_{t}) \\ f^{k}(q^{k}_{t}, x_{t})
\end{bmatrix}
\end{aligned}
$}
\label{nonlinear_statespace}
\end{equation}
\\
where $\mathbb{1}$ is the $n \times n$ identity matrix. The system in Equation~\eqref{nonlinear_statespace} is a nonlinear time-varying dynamical system, which makes it difficult to analyze whether or not the states $q^{n}_{t}$ for $n=1,\hdots,k$ are prone to diverge. However, by Lyapunov's linearization method, the system can be linearized to study the stability of the states in a region around the equilibrium points $q^{e}$ and $x^{e}$. The equilibrium points can be easily found by setting $q_{t+1} = f(q^{e},x^{e}) = 0$, which vary according to the chosen activation function. The linearized system can be found by taking the Jacobian of the functions $f(q_{t}, x_{t})$ with respect to the states and input to the system evaluated at the equilibrium points, which has the form: 

\begin{equation}
\resizebox{0.45\textwidth}{!}{$
q_{t+1} = \evalat[\big]{\nabla_{q}f(q_{t},x_{t})}{\overset{q_{t} = q^{e}_{t}}{x_{t}=x^{e}_{t}}}q_{t} + \evalat[\big]{\nabla_{x}f(q_{t},x_{t})}{\overset{q_{t} = q^{e}_{t}}{x_{t}=x^{e}_{t}}}x_{t} + h.o.t. 
$}
\end{equation}
\\
where $h.o.t.$ are the higher order terms of the true nonlinear system. In matrix form the linearized system, after neglecting the higher order terms, is written as:

\begin{equation}
\resizebox{0.45\textwidth}{!}{$
\begin{bmatrix}
q_{t+1}^{1} \\ q_{t+1}^{2} \\ \vdots \\ q_{t+1}^{k-1} \\ q_{t+1}^{k}
\end{bmatrix}
\approx
\underbrace{
\evalat[\Bigg]{\begin{bmatrix}
\alpha_{1} + \frac{\partial \varphi(\cdot)}{\partial q^{1}_{t}} & \alpha_{2} & \hdots & \alpha_{k-1} & \alpha_{k} \\
\mathbb{1} & \mathbb{0} & \hdots & \mathbb{0} & \mathbb{0} \\
\mathbb{0} & \mathbb{1} & \hdots & \mathbb{0} & \mathbb{0} \\
\vdots & \ddots  & \ddots & \vdots & \vdots \\
\mathbb{0} & \hdots & \mathbb{0} & \mathbb{1} & \mathbb{0} \\
\end{bmatrix}}{\overset{q_{t} = q^{e}_{t}}{x_{t}=x^{e}_{t}}}}_{A_{(n \cdot k \times n \cdot k)}}
\begin{bmatrix}
q_{t}^{1} \\ q_{t}^{2} \\ \vdots \\ q_{t}^{k-1} \\ q_{t}^{k}
\end{bmatrix}
+
\underbrace{
\evalat[\Bigg]{\begin{bmatrix}
\frac{\partial \varphi(\cdot)}{\partial x_{t}} \\ \mathbb{0} \\ \vdots \\ \mathbb{0} \\ \mathbb{0}
\end{bmatrix}}{\overset{q_{t} = q^{e}_{t}}{x_{t}=x^{e}_{t}}}}_{B_{(n \cdot k \times 1)}}
x_{t} 
$}
\label{linearized_matrixform}
\end{equation}
\\
The assumption made here is that the activation function used is continuously differentiable and locally Lipschitz. The activation function chosen for our experiments is the hyperbolic tangent function, $\tanh( \cdot )$, which is continuously differentiable and locally Lipschitz around the origin. We know $\tanh(q^{1}_{t}, x_{t})=0$ when $(q^{1}_{t},x_{t})=(0,0)$ for all $t$, and $\evalat{\frac{\partial \varphi (\cdot)}{\partial q_{t}^{1}}}{\overset{q_{t} = 0}{x_{t}=0}} = W_{rec}$, where $W_{rec}$ is the recurrent weight matrix. By setting all of the state functions $f(q_{t}^{e}, x_{t}^{e}) = 0$, all of the resulting equilibrium points will be at the origin.

From Lyapunov's linearization method, it is possible to determine whether the states $q^{j}_{t}$ will converge or not $\forall j = 1,2, \dots, k$, by studying the eigenvalues, $\lambda(A)$, of the matrix $A$ in equation \eqref{linearized_matrixform} above. Since the system is discrete, to ensure convergence of the states, we require the absolute value of all of the eigenvalues of $A$ to be strictly less than one~\cite{krstic1995nonlinear}. In other words, the desired eigenvalues should be inside the unit circle:

\begin{equation}
    |\lambda_{i}^{*}| < 1, \forall i = 1, \hdots, n \cdot k
\label{des_eig}
\end{equation}
where $\lambda_{i}^{*}$ is the $i^{th}$ desired eigenvalue.

\subsection{Regularizer for Desired Eigenvalues}
As mentioned earlier, the linearized model of the system varies at every training iteration, thus the skip-coefficients constantly need to be adjusted in order to maintain all desired eigenvalues, $\lambda^{*}_{i}$ for $i = 1, \hdots, n \cdot k$, of the system. Since the skip-coefficients of the network are subject to change at every training iteration, the loss function with respect to the learnable parameters of the network will also change at every training iteration. 

For this reason, the parameters for the skip-coefficients are also added as an eigenvalue regularization term in the loss function. In this way the parameter updates take into consideration both the eigenvalue placement, as well as minimizing the classification/regression error.

The eigenvalue regularizer we built is designed to minimize the distance between  $\lambda^{*}_{i}$ for $i = 1, \hdots, n \cdot k$ and the eigenvalues of $A$ in \eqref{linearized_matrixform} as a function of the skip-coefficients, $\lambda(\alpha)$, by using the Euclidean norm:

\begin{equation}
\begin{aligned}
    C_{\lambda}(\lambda^{*}, \lambda(\alpha)) = \sqrt{\sum_{i = 1}^{n \cdot k}{\big(\lambda_{i}^{*} - \lambda_{i}(\alpha)\big)^2}}
\end{aligned}
\end{equation}
\\
Note that the complexity for computing the eigenvalues of a square $m \times m$ matrix increases by the order of $\mathcal{O}(m^{2})$. If we denote the cost-function for output of the neural network as $C_{NN}$, then the overall cost function of the \rnn becomes:
\begin{equation}\label{eq:total_loss}
    C_{\rnn} = C_{NN} + \beta\cdot C_{\lambda}
\end{equation}
where $\beta$ is a constant regularizer parameter and can be simply chosen to be $1$ as in our experiments.


\paragraph{\textit{Remark} 1.} The skip-coefficients and the rest of the parameters of the \rnn are updated simulltaneously, which indicates that the skip-coefficients no longer yield the desired eigenvalues of the new linearized model. However, this does not pose an alarming problem if the learning rate is chosen small enough. This is because the parameters will be updated slow enough such that the new set of $\lambda(A^{(t+1)})$ are in a region close enough to $\lambda(A^{(t)})$, where the superscript $(t)$ refers to the training iteration. \qed

\paragraph{\textit{Remark} 2.} From \eqref{des_eig} all the desired eigenvalues must belong inside the unit circle. The closer the eigenvalues are chosen to the origin, the faster the states should converge towards the corresponding equilibrium points, whereas eigenvalues chosen very close to $1$ will require a large number of time-steps for $h_{t}$ to converge. 
Hence, intuitively one would choose a small desired eigenvalue to ensure a more stable trajectory of hidden states, which effectively allows the RNN to learn better. \qed

\subsection{Backpropagation Through Time with \rnn}

Backpropagation through time (BPTT) is well understood and established \cite{werbos1990backpropagation}, especially for a standard RNN with no skip connections. However, when skip-connections are introduced into the architecture, as in the \rnn, the derivation of the BPTT becomes drastically more complex to expand as a function of time. This is because the chain-rule adapted in the standard BPTT needs to be reapplied to all $k$ previously connected hidden states. In equation form, this expands to include multiple nested summation terms which have the following structure:

\begin{equation}
\resizebox{0.48\textwidth}{!}{$
\begin{aligned}
    \frac{\partial \mathcal{E}_{t}}{\partial h_{t-T}} &=
    \frac{\partial \mathcal{E}_{t}}{\partial h_{t}}
    \Bigg(
    \sum_{i^{(1)}=1}^{k}{\frac{\partial h_{t}}{\partial h_{t-i^{(1)}}}}
    \Bigg) \\
    &= \frac{\partial \mathcal{E}_{t}}{\partial h_{t}}
    \Bigg(
    \sum_{i^{(1)}=1}^{k}{\frac{\partial h_{t}}{\partial h_{t-i^{(1)}}}}
    \Bigg(
    \sum_{i^{(2)}=1}^{k}{\frac{\partial h_{t-i^{(1)}}}{\partial h_{t-i^{(1)}-i^{(2)}}}}
    \Bigg)
    \Bigg) \\
    &= \frac{\partial \mathcal{E}_{t}}{\partial h_{t}}
    \Bigg(
    \sum_{i^{(1)}=1}^{k}{\frac{\partial h_{t}}{\partial h_{t-i^{(1)}}}}
    \Bigg(
    \sum_{i^{(2)}=1}^{k}{\frac{\partial h_{t-i^{(1)}}}{\partial h_{t-i^{(1)}-i^{(2)}}}}
    \Bigg(
    \sum_{i^{(3)}=1}^{k}{\frac{\partial h_{t-i^{(1)}-i^{(2)}}}{\partial h_{t-i^{(1)}-i^{(2)}-i^{(3)}}}}
    \Bigg)
    \Bigg)
    \Bigg)
    \\
    &\vdots \\
    &= 
    \frac{\partial \mathcal{E}_{t}}{\partial h_{t}}
    \Bigg(
    \sum_{i^{(1)}=1}^{k}{\frac{\partial h_{t}}{\partial h_{t-i^{(1)}}}}
    \Bigg(
    \sum_{i^{(2)}=1}^{k}{\frac{\partial h_{t-i^{(1)}}}{\partial h_{t-i^{(1)}-i^{(2)}}}}
    \Bigg(
    \hdots 
    \Bigg(
    \sum_{i^{(T)}=1}^{k}{\frac{\partial h_{t-i^{(1)}-\hdots -i^{(T-1)}}}{\partial h_{t-i^{(1)}-\hdots-i^{(T)}}}}
    \Bigg)
    \Bigg)
    \Bigg)
    \Bigg)
\end{aligned}
$}
\label{DCRNN_BPTT}
\end{equation}
where $T \geq k$. This expansion can prove to be tedious at times, especially for larger values of $k$; thus, we present an alternative approach to expand Equation~\eqref{DCRNN_BPTT} in the Supplementary Materials. Once the equation is expanded to the $T^{th}$ time step, the following substitutions are recommended:

\begin{equation}
\begin{aligned}
    \frac{\partial h_{t}}{\partial h_{t-1}} &= \alpha_{1} + W_{rec}^{T}diag\big(\varphi^{'}(h_{t})\big) \\
    \frac{\partial h_{t}}{\partial h_{t-i}} &= \alpha_{i} \hspace{10mm} 1 < i \leq k
\end{aligned}
\label{BPTT_subs}
\end{equation}

We present two examples to illustrate BPTT for the \rnn for the cases where $k=1$ and $2$, respectively:

\begin{enumerate}
    \item For the $k=1$ case, we show the effect of the BPTT on the gradient by borrowing notations from the proof in \cite{pascanu2013difficulty}, as follows:

We know that the Jacobian matrix $\frac{\partial h_{t}}{\partial h_{t-1}} = \alpha_{1} + W^{T}_{rec}diag\big(\varphi^{'}(h_{t-1})\big)$.  Let $\Lambda_{w} = \frac{1}{\gamma}$ be the largest singular value of the recurrent weight matrix $W_{rec}$, and $\Lambda_{\alpha}$ be the largest singular value of the skip-coefficient $\alpha_{1}$. Then $\forall t$:

\begin{equation}
\resizebox{0.45\textwidth}{!}{$
    \norm{\frac{\partial h_{t+1}}{\partial h_{t}}} 
    \leq 
    \norm{\alpha_{1}} + \norm{W^{T}_{rec}} \norm{diag\big(\varphi^{'}(h_{t})\big)} 
    \leq 
    \Lambda_{\alpha} + \frac{1}{\gamma}\gamma \leq \Lambda_{\alpha} + 1
    $}
\end{equation}
\\
Let $\eta \in \mathbb{R}$ be such that $\norm{W^{T}_{rec}} \norm{diag\big(\varphi^{'}(h_{t})\big)} = \eta$. Let the error of the RNN at time $t$ be denoted by $\mathcal{E}_{t}$. Then it can be shown that
\begin{equation}
    \norm{\frac{\partial \mathcal{E}_{t}}{\partial h_{t}} \Bigg(\prod_{i=T}^{t-1} \frac{\partial h_{i+1}}{\partial h_{i}} \Bigg)} 
    \leq 
    (\Lambda_{\alpha}  + \eta)^{t-T} \norm{\frac{\partial \mathcal{E}_{t}}{\partial h_{t}}} 
\end{equation}


It can be seen that the \rnn for $k=1$ case may still suffer from the vanishing gradients problem. However, as mentioned earlier, the \rnn is not designed to solve memory-based tasks, but rather to learn the underlying equations that drive dynamic systems. The long-term dependencies problem applies to memory-based problems and not dynamics-based problems, which will be discussed in detail in Section~\ref{sec:exp_class}.

\item For the $k=2$ case, we give an example of BPTT for $T=6$. 


Using equations \eqref{DCRNN_BPTT} and \eqref{BPTT_subs}, and defining $\alpha_{1} + W_{rec}^{T}diag\big(\varphi^{'}(h_{t-1})\big) \triangleq \alpha_{1}^{*}$, the example unfolds to be:

\begin{equation}
\resizebox{0.44\textwidth}{!}{$
\begin{aligned}
    \frac{\partial \mathcal{E}_{t}}{\partial h_{t-6}} &= \frac{\partial \mathcal{E}_{t}}{\partial h_{t}} \Bigg( \frac{\partial h_{t}}{\partial h_{t-1}} + \frac{\partial h_{t}}{\partial h_{t-2}}
    \Bigg)
    \\
    &= \frac{\partial \mathcal{E}_{t}}{\partial h_{t}} \Bigg( \frac{\partial h_{t}}{\partial h_{t-1}}
    \Bigg( \frac{\partial h_{t-1}}{\partial h_{t-2}} + \frac{\partial h_{t-1}}{\partial h_{t-3}}
    \Bigg)
    + \frac{\partial h_{t}}{\partial h_{t-2}}
    \Bigg( \frac{\partial h_{t-2}}{\partial h_{t-3}} + \frac{\partial h_{t-2}}{\partial h_{t-4}}
    \Bigg)
    \Bigg) \\
    \vdots \\
    &=\Big(\frac{\partial h_{t}}{\partial h_{t-1}}\Big)^{6}
    + 
    5\Big(\frac{\partial h_{t}}{\partial h_{t-1}}\Big)^{4}\Big(\frac{\partial h_{t}}{\partial h_{t-2}}\Big)^{1}
    \\
    & \hspace{58pt} + 6\Big(\frac{\partial h_{t}}{\partial h_{t-1}}\Big)^{2}\Big(\frac{\partial h_{t}}{\partial h_{t-2}}\Big)^{2}
    + 
    \Big(\frac{\partial h_{t}}{\partial h_{t-2}}\Big)^{3} \\
    & = 
    (\alpha_{1}^{*})^{6} 
    + 
    5(\alpha_{1}^{*})^{4}(\alpha_{2})^{1}
    +
    6(\alpha_{1}^{*})^{2}(\alpha_{2})^{2}
    +
    (\alpha_{2})^{3}
\end{aligned}
$}
\label{k2_example}
\end{equation}

\end{enumerate}

\section{Experiments and Results}
In this section, we test the performance of the proposed \rnn on several different tasks. We compare the results with (i) a vanilla RNN with tanh activations, as well as (ii) a LSTM. As previously stated, the explicit hypothesis that we are testing is that memory-based tasks and dynamics-based tasks are two very different kinds of tasks, and as powerful a model as the LSTM is, it seems to be ill-suited for modeling dynamical systems. First, a numerical dataset was created to test the forecasting performance of the \rnn for data that is generated by a set of ODEs. Second, we test the performance of the \rnn for a classification task, as well as how different $k$ parameters in the \rnn leads to different results. We also explain why the long-term dependencies problem does not apply to dynamical problems. Third, we test the performance on the MNIST dataset and a copy memory task \cite{arjovsky2016unitary}).

For all models, we used a single layer with $128$ hidden units and a batch size of $1000$. All of the networks were trained with the Adam optimizer \cite{kingma2014adam} with a learning rate of $0.001$,  with gradients clipped to $5$~\cite{pascanu2013difficulty}. For \rnn models, we initialized the weight matrices with Glorot initializers~\cite{glorot2010understanding}. All models have been implemented in Tensorflow~\cite{abadi2016tensorflow}.

\subsection{Lorentz System Forecasting}

Standard machine learning datasets mostly belong to the domains of images or speech, where the sequential data are often not driven by an underlying differential equation. In other words, most of these standard tasks are memory-based tasks and are not dynamics-based tasks. Because of this, we created a simple numerical example to test the performance of the \rnn with other standard methods.

\begin{figure}{\vspace{0pt}}  
  \centering
  \includegraphics[width=0.8\columnwidth]{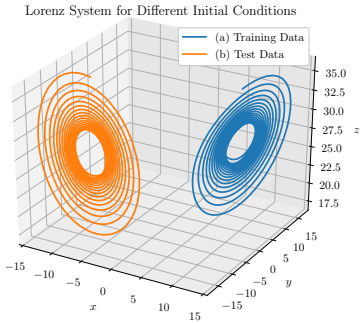}
  \caption{Example of Lorenz systems with parameter $\sigma=10, \rho=28, \beta=8/3$ and different initial conditions. For training data, the initial conditions are (a) $x=6$, $y=8$, $z=25$ and for test data, the initial conditions are (b) $x=-6$, $y=-8$, $z=25$. It is seen that in order to correctly forecast a test trajectory from the training trajectory, it is necessary for the network to learn the underlying differential equation governing the evolution of the training trajectory, instead of trying to remember what the training trajectory looks like.}
  \label{fig_lorenz_ic}
  \hspace{-1 pt}
\end{figure}

The Lorentz system is a chaotic dynamical system governed by three coupled differential equations~(\ref{eq:lorenz}).
\begin{align}\label{eq:lorenz}
&    \frac{dx}{dt}=\sigma(y-x) \nonumber\\
&    \frac{dy}{dt}=x(\rho-z)-y\\
&    \frac{dz}{dt}=xy-\beta z \nonumber
\end{align}
Even though the system is chaotic (and so a slight change in initial conditions can yield huge changes in the trajectory), if a neural network is able to learn the underlying differential equation governing the system, then given a new initial condition the same learned neural network should still be able to well model this system. On the other hand, if the neural network is simply trying to remember what the trajectories look like without capturing the equations' underlying dynamics, it will not perform well at this task. See Figure~\ref{fig_lorenz_ic} for an example of two trajectories generated by the exact same system but with two different initial conditions.

This hypothesis was tested with a typical Lorenz System of $\sigma=10$, $\rho=28$ and $\beta=8/3$, which yield the well known figure-eight-like trajectory. A collection of $200$ initial conditions were randomly sampled from a 3-dimensional Gaussian (since the Lorenz System's feature dimension is 3 for the ($x,y,z$) initial conditions) with zero mean and standard deviation $10$. Half of the initial conditions were used to produce a training trajectory of $1,000$ samples and a time step of $T=10$ for each of the samples, while the other half were used to produce a testing trajectory of the same length. The goal of the neural network is to forecast the next time step given the present as $ x_{t+1} = x_{t} + \frac{df\left(x_t,t\right)}{dt}\Delta t$, where $\Delta t$ is set to be $0.01$ for all cases. This is thus a direct means of measuring the ability for the neural network to learn the underlying model for the Lorentz system, as the training and testing trajectories can be wildly different due to the chaotic dynamics (as illustrated by example in Figure~\ref{fig_lorenz_ic}) even though the underlying dynamical system is the same. The error is measured as the Euclidean distance between the true position and the predicted position, i.e. $\textit{E}_{t+1}:=||x_{t+1} - \hat{x}_{t+1} ||_2$.

The results of these experiments are listed in Table~\ref{tab:forecasting}, where it is seen that the dynamical systems model outperforms the standard RNN model $100$ out of $100$ experiments. Additionally it is seen that the memory based LSTM model is ill-suited for this task of learning a dynamical system, as it performs the least well in all $100$ experiments.

\begin{table}[h]
\centering
  \begin{tabular}{  l | c | c | r }
    \toprule
    & $1^{st}$ & $2^{nd}$ & $3^{rd}$ \\ 
    \midrule
    \rnn & \textbf{100} & $0$ & $0$ \\ 
    RNN & $0$ & $100$ & $0$ \\ 
    LSTM & $0$ & $0$ & $100$ \\
    \bottomrule
  \end{tabular}
  \caption{Forecasting results on the Lorentz system with the best result in bold. This shows that the dynamical systems model outperforms the standard RNN model $100$ out of $100$ randomized experiments. Additionally it is seen that the memory based LSTM model is ill-suited for this task of learning a dynamical system.}
  \label{tab:forecasting}
\end{table}


When comparing the reductions in error on this task, averaged over the $100$ randomized trials, the \rnn reduces the error of the LSTM by $80.0\% \pm 2.99\%$, and the \rnn reduces the error of the RNN by $62.0\% \pm 5.92\%$.

These results show that the LSTM, although extremely powerful when used to model systems that have some internal memory, has significant difficulty in modeling dynamical systems of differential equations, as its structure does not lend itself well to this task. Conversely, the \rnn was specifically designed to be able to model dynamical systems, which is why it has the ability to excel at this task.

\color{black}

\subsection{Lorentz System Classification}\label{sec:exp_class}



This subsections discusses how the parameter $k$ in \rnn affects the model accuracies, and also discusses why the long-term dependencies problem does not apply to dynamics-based tasks.

\begin{figure}{\vspace{0pt}}  
  \includegraphics[width=\columnwidth]{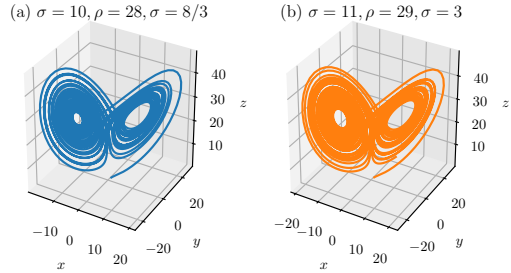}
  \caption{Lorenz systems with $2$ sets of parameters (a) $\sigma=10, \rho=28, \beta=8/3$ and (b) $\sigma=11, \rho=29, \beta=3$.}
  \label{fig_lorenz}
\end{figure}

For this experiment, we created two Lorenz systems with parameters (a) $\sigma=10, \rho=28, \beta=8/3$ and (b) $\sigma=11, \rho=29, \beta=3$ for classification. The samples from both classes were generated with the same initial conditions of $x=1, y=1, z=1$ and integration step of $\Delta t=0.01$. The lengths of the time steps $T$ consist of these following values: $T\in \{10,20,50,100,200,500\}$. Starting at the initial conditions, $T$ time steps of $\{x,y,z\}$ values will be generated as one sample in the dataset. Then we repeat $20,000$ times for each $T$ value to get $10,000$ training data (randomly chosen from all samples) and the rest as test data. The two systems' trajectories are plotted in Figure~\ref{fig_lorenz}.

In Table~\ref{table:model_k} we list the accuracies for different values of $k$ parameter in the \rnn as well as for vanilla RNN and LSTM when $T=10$. It is shown that as that \rnn with $k=1$ will yield the overall best performance for this classification task. It is consistent with the Lorenz System equations~(\ref{eq:lorenz}) since the current state only depend on its one time-step previous state. Enforcing additional structures on the hidden states by adding hidden states further back in history will lead to worse results. Additionally, as previously discussed in Section~\ref{sec:propose}, a LSTM is a $k=1$ network with additional gating functions, and a vanilla RNN is a $k=0$ network. The proposed \rnn with $k=1$ can outperform both networks. 

\begin{table}[h]
\centering
\caption{Lorenz System classification training and test accuracies for \rnn with $k=1,2,3,4,5$, vanilla RNN and LSTM. The time steps (sequence length) for each of the sample is $T=10$. The means and standard deviations from $10$ independent runs are computed. Highest test set accuracy is presented in bold. All models are trained with $128$ hidden units and $1000$ epochs.}
\scalebox{1.0}{
\begin{tabular}{c|c|c}
\toprule
\textbf{Models} & \textbf{Training Set} & \textbf{Test Set}  \\
\midrule
 \rnn $k=1$  &  0.9215$\pm$0.0342  & \textbf{0.9399}$\pm$0.0131  \\ 
 \rnn $k=2$  &  0.9266$\pm$0.0313  &  0.9287$\pm$0.0297  \\
 \rnn $k=3$  &  0.9202$\pm$0.0539  &  0.9209$\pm$0.0488  \\
 \rnn $k=4$  &  0.9378$\pm$0.0250  &  0.9143$\pm$0.0201  \\
 \rnn $k=5$  &  0.9438$\pm$0.0258  &  0.9115$\pm$0.0240  \\
 RNN   &  0.9618$\pm$0.0136  &  0.8916$\pm$0.0212  \\
 LSTM  &  0.9837$\pm$0.0075  &  0.7164$\pm$0.0465  \\
\bottomrule
\end{tabular}
}
\vspace{0pt}
\label{table:model_k}
\vspace{0pt}
\end{table}

\begin{figure}{\vspace{0pt}}  
  \includegraphics[width=\columnwidth]{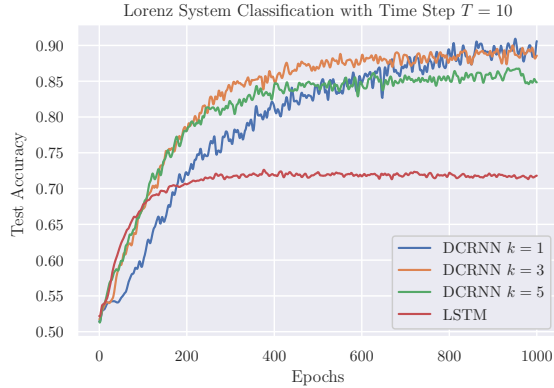}
  \caption{Lorenz systems classification of two systems (a) $\sigma=10, \rho=28, \beta=8/3$ and (b) $\sigma=11, \rho=29, \beta=3$, test accuracy plots for the \rnn with $k=1,3,5$ and LSTM versus number of epochs. The average of $10$ independent runs are reported. It can be easily seen that the tasks becomes easier as $T$ increases.}
  \label{fig_timestep_10}
\end{figure}

Figure~\ref{fig_timestep_10} shows the average test set accuracy for $10$ independent training sessions versus the number of epochs. Specifically, the \rnn with $k=1,3,5$ and the LSTM results are shown. As the parameter $k$ increases, the \rnn requires less epochs to converge as the network is doing more computations (adding $k$-step previous hidden states and calculating eigenvalues of a larger square matrix, the latter of which is much more computationally expensive) per epoch. It is seen that the LSTM is unable to properly learn the dynamical structure of this process, and is vastly overfitting the data.

\begin{figure}{\vspace{0pt}}  
  \includegraphics[width=\columnwidth]{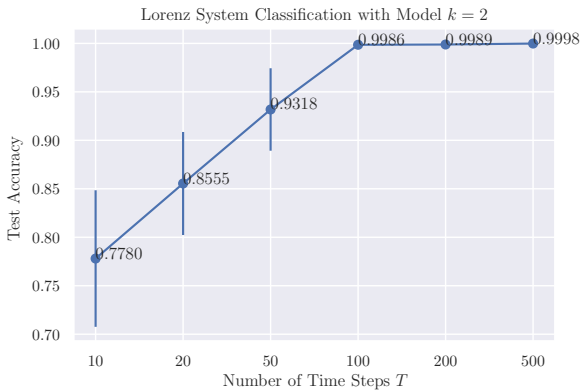}
  \caption{Lorenz System classification with the same \rnn model of parameter $k=2$ and varying time steps $T$ in the dataset for $200$ training epochs. All models are trained $10$ times independently with their average test accuracies plotted and annotated in the figure, and their standard deviations shown as the error bars in the figure.}
  \label{fig_k_2}
\end{figure}

Furthermore, we will try to illustrate why the so-called ``long term dependency problem''~\cite{bengio1994learning} is not applicable to dynamics-based tasks. Figure~\ref{fig_k_2} shows the average test accuracy and standard deviation for different time step length $T$ for the \rnn with parameter $k=2$. Contrary to many sequential tasks where the problem becomes harder with increasing sequence length, this classification task becomes easier with larger $T$, as seen in Figure~\ref{fig_k_2}. A simple geometric interpretation is that we are taking a longer curve of the trajectory generated by the underlying ODEs when we are increasing the sequence length, and the longer the curve is, the easier it is to classify. Therefore, we believe that ``sequential tasks'' should be carefully divided into two subcategories: memory-based sequential tasks (e.g. speech data) and dynamics-based sequential tasks (e.g. engineering systems sensory data), due to the vast differences in their behaviours with long-term dependencies. Methods that work extremely well for one of those tasks (e.g. LSTM for memory-based tasks) can easily fail at the other tasks (e.g. LSTM for forecasting or classification of Lorenz Systems), and vice versa.

\subsection{MNIST and Copying Memory Problem}
Although our proposed \rnn is not designed for memory-based tasks, we still present the results on MNIST and the copying memory problem~\cite{arjovsky2016unitary}. 

For row-by-row MNIST, we treat the $28\times28$ images as sequential data of $28$ time steps with $28$-dimensional features at each time step. Test accuracies after $200$ epochs are reported in Table~\ref{table:mnist}. The LSTM performs better than the \rnn. Since there is no dynamical relationship between two adjacent time steps in this dataset, i.e. sequential MNIST was not generated by a system of differential equations, the \rnn is not able to outperform the LSTM with its memory structure.

\begin{table}[h]
\centering
\caption{Row-by-row MNIST classification for \rnn with $k=1,2,3,4,5$, vanilla RNN and LSTM. The means and standard deviations from $10$ independent runs are computed. Highest test set accuracy is presented in bold. All models are trained with $128$ hidden units and $200$ epochs.}
\scalebox{1.0}{
\begin{tabular}{c|c}
\toprule
\textbf{Models} & \textbf{Test Set Accuracy}  \\
\midrule
 \rnn $k=1$  &  0.9703$\pm$0.0038  \\ 
 \rnn $k=2$  &  0.9726$\pm$0.0026  \\
 \rnn $k=3$  &  0.9732$\pm$0.0023  \\
 \rnn $k=4$  &  0.9736$\pm$0.0022  \\
 \rnn $k=5$  &  0.9745$\pm$0.0017  \\
 RNN   &  0.9780$\pm$0.0020  \\
 LSTM  &  \textbf{0.9849}$\pm$0.0011  \\
\bottomrule
\end{tabular}
}
\vspace{0pt}
\label{table:mnist}
\vspace{0pt}
\end{table}

\begin{figure}{\vspace{0pt}}  
  \includegraphics[width=\columnwidth]{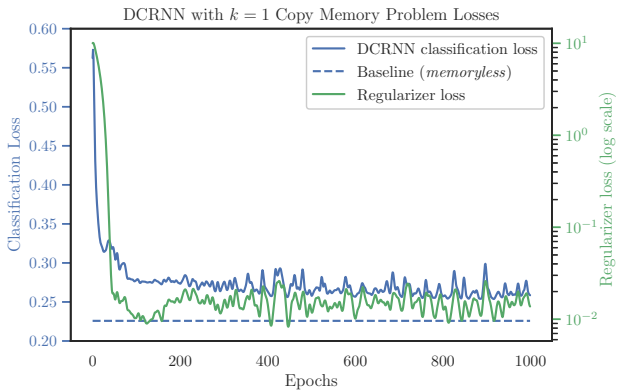}
  \caption{Loss functions for the test data for copying memory problem for time lag $T=20$. The blue solid line is the cross entropy loss for the classification results, and the green solid line is the \rnn regularizer loss. The blue dashed line represents the baseline classification loss for a memoryless strategy. Note that the total loss for optimizing the \rnn is the sum of the classification loss and regularizer loss as in Equation~(\ref{eq:total_loss}).}
  \label{fig:copy_mem}
\end{figure}

The copying memory problem is a more direct way of testing if a model has an explicit memory, and it considers $10$ categories $\{a_i\}^9_{i=0}$ of $T+20$ length data. In the input, the first $10$ entries contains actual information that needs to be remembered, and are sampled uniformly from $\{a_i\}^7_{i=0}$. The next $T-1$ entries are all $a_8$, followed by a delimiter $a_9$. The output consists of $T+10$ repeated entries of $a_8$ followed by the first $10$ entries from the input. This is a pure memory problem and the goal is to minimize the cross entropy of the predictions at each time step. We can plot the losses for the case of $T=20$ in Figure~\ref{fig:copy_mem}.

For this purely memory-based task, it has been shown that a baseline memoryless strategy (which would predict $a_8$ for $T+10$ entries and random guess among $\{a_i\}_{i=0}^7$ for last $10$ entries) would lead to a categorical loss of $\frac{10\log(8)}{T+20}$. To beat this baseline, the system need to be able to process memory, which is why the \rnn has a higher loss than the baseline. As good as the \rnn is for modeling dynamical systems governened by ODEs, it is not suitable for memory-based tasks because it does not have a memory structure.

\section{Summary and Discussions}
This paper develops a new RNN architecture designed to model data generated by dynamical systems that are governed by ODEs. The \rnn is a generalization of the vanilla RNN, in that it allows for weighted skip connections to the current hidden state across $k$-step previous hidden states. From the perspective of dynamical systems and control theory, the notion of Lyapunov stability is introduced to an eigenvalue regularization term in the overall loss function, which imposes a controller for the hidden state trajectory. Experimental results show that the \rnn is able to outperform LSTM in forecasting and classification tasks of dynamical systems by a substantial margin.

One of the notable findings in the experiments with the \rnn is that ``sequential data'' can have very different underlying structures depending on how the data was generated. Data generated by dynamical systems in many physics and engineering domains are often governed by some system of differential equations, while this is not the case for speech or image data~\cite{chomsky2002syntactic}. It follows that we should treat memory-based tasks and dynamics-based tasks differently, instead of trying to use a ``one-size-fits-all'' approach for all sequential time series data. 

It was shown that the memory structure in the LSTM, which makes it so effective for memory-based tasks, is also it's own Achilles' heel when modeling dynamical systems, because trying to remember trajectories without learning the dynamical structure is ineffective for machine learning tasks involving nonlinear differential equations. What's more, there is a need for developing standard sequential datasets for dynamical systems. We believe that the MNIST dataset is inappropriate for evaluating sequential models, yet for the lack of standard preprocessed sequential datasets, it is still being used fairly often for proof of concepts. 

There are many directions of future research in further developing the \rnn. It is obvious that if we know the data is generated by a $k$-th order dynamical system, we should use the same $k$ parameter in the \rnn. However, a systematic method for selecting the value of the $k$ parameter, given a dataset with unknown dynamics, is an important direction for future study. Additionally, a more rigorous study for selecting the eigenvalues will also help improve the model. Finally, studies on intializing the weight matrix could potentially make the training converge faster. 




\section*{Supplementary Material}

\subsection*{Backpropagation Through Time: Combinatorics Shortcuts}
The expansion of Backpropagation Through Time (BPTT) for the DCRNN, introduced in Section 3.3, may seem tedious to expand, especially for larger values of $k$. We present methods using combinatorics to simplify the calculations relating to the expansion of the BPTT.

Due to the space constraint, we will show the example of $k=2$ and $T=4$ here. This can be easily generalized to multiple time-steps and different $k$ values. For this case, the DCRNN has the following structure: $ h_{t} = \alpha_{1}h_{t-1} + \alpha_{2}h_{t-2} + \varphi_h(h_{t-1},x_{t})
$.
By defining the partial derivatives $\frac{\partial h_{t}}{\partial h_{t-1}} = \alpha_{1} + W_{rec}^{T}diag\big(\varphi^{'}(h_{t-1})\big) \triangleq \alpha_{1}^{*}$ and $\frac{\partial h_{t}}{\partial h_{t-2}} \triangleq \alpha_{2}$, we can draw the following tree:

\vspace{-10pt}
\begin{center}
\begin{forest}
[$h_t$ 
    [$h_{t-1}$ , edge label={node[midway,left,font=\scriptsize] {$\alpha_{1}^{*}$}}
      [$h_{t-2}$, edge label={node[midway,left,font=\scriptsize] {$\alpha_{1}^{*}$}} 
        [$h_{t-3}$, edge label={node[midway,left,font=\scriptsize] {$\alpha_{1}^{*}$}} 
            [$h_{t-4}$ , edge label={node[midway,left,font=\scriptsize] {$\alpha_{1}^{*}$}}],
            [,phantom]
        ]
        [$h_{t-4}$, edge label={node[midway,right,font=\scriptsize] {$\alpha_{2}$}}]
      ] 
      [$h_{t-3}$, edge label={node[midway,right,font=\scriptsize] {$\alpha_{2}$}}
        [$h_{t-4}$, edge label={node[midway,left,font=\scriptsize] {$\alpha_{1}^{*}$}}]
        [,phantom]
      ] 
    ]
    [$h_{t-2}$, edge label={node[midway,right,font=\scriptsize] {$\alpha_{2}$}}
      [$h_{t-3}$, edge label={node[midway,left,font=\scriptsize] {$\alpha_{1}^{*}$}}
        [$h_{t-4}$, edge label={node[midway,left,font=\scriptsize] {$\alpha_{1}^{*}$}}]
        [,phantom]
      ] 
      [$h_{t-4}$, edge label={node[midway,right,font=\scriptsize] {$\alpha_{2}$}}
      ]
  ] 
]
\end{forest}
\vspace{-10pt}
\end{center}

Starting at $h_t$, where $h_t=f(h_{t-1},h_{t-2})$, the error being backpropagated will split to both $h_{t-1}$ (left child) and $h_{t-2}$ (right child), with the edge of the tree denoting the respected partial derivatives. The error can be traced back through several branches until $h_{t-4}$ is reached. By summing up all the branches we should get the total error propagated to $T=4$ time steps back. $\frac{\partial \mathcal{E}_{t}}{\partial h_{t-4}} = (\alpha_{1}^{*})^{4} +       3(\alpha_{1}^{*})^{2}(\alpha_{2})^{1} + (\alpha_{2})^{2}$. Note, the value of $k$ dictates the number of branches each edge splits into, meaning for $k=1$ trees or $k=0$ trees (vanilla RNN), they will only have the left child because the current hidden state only depend on its immediate previous hidden state, i.e. $h_t=f(h_{t-1})$.

Now for this case ($k=2$ and $T=4$), we can alternatively treat this problem as trying to go back $4$ time steps in total, when you can go back $1$ step $(\partial h_t/\partial h_{t-1})$ or $2$ steps $(\partial h_t/\partial h_{t-2})$ at a time. Then, this becomes a problem of finding the pair of natural numbers $(i,j)$ such that
\begin{equation}
    \evalat[\big]{T = i \times 1 + j \times 2}{i,j \in \mathbb{N}}
  \label{comb_k2} 
\end{equation}    
If $T=4$, then we have $4=4\times1=2\times1+1\times2=2\times2$, or $(i,j)\in((4,0),(2,1),(0,2))$. Looking back at the tree graph, since going left is $1$ step and going right is $2$ steps, we can take $4$ left steps, or $2$ left steps and $1$ right step, or $2$ right steps to reach $h_{t-4}$. Once we've settled for each pair of $(i,j)$ values, the associated term is given by
\begin{equation}
   \binom{i+j}{i}\Bigg(\frac{\partial h_{t}}{\partial h_{t-1}}\Bigg)^{i} \cdot \Bigg(\frac{\partial h_{t}}{\partial h_{t-2}}\Bigg)^{j}
  \label{coeff} 
\end{equation} 
For example, when $(i,j)=(2,1)$, by using combinatorics we can easily determine that coefficient for the term $(\alpha_{1}^{*})^{2}(\alpha_{2})^{1}$ is $\binom{2+1}{2}=3$ (basically we are taking $2$ left children and $1$ right child from the tree, and we need to choose the locations of the two left children among all three children). 

Then for the derivation of $\frac{\partial \mathcal{E}_{t}}{\partial h_{t-6}}$ for a DCRNN of parameter $k=2$ in Section 3.3, we can write the BPTT without having to expand every term by

\vspace{-10pt}
\begin{equation}
\begin{aligned}
    \frac{\partial \mathcal{E}_{t}}{\partial h_{t-6}} 
    &=\binom{6}{6}\Big(\frac{\partial h_{t}}{\partial h_{t-1}}\Big)^{6}
    + 
    \binom{5}{4}\Big(\frac{\partial h_{t}}{\partial h_{t-1}}\Big)^{4}\Big(\frac{\partial h_{t}}{\partial h_{t-2}}\Big)^{1}
    \\
    &+ \binom{4}{2}\Big(\frac{\partial h_{t}}{\partial h_{t-1}}\Big)^{2}\Big(\frac{\partial h_{t}}{\partial h_{t-2}}\Big)^{2}
    + 
    \binom{3}{3}\Big(\frac{\partial h_{t}}{\partial h_{t-2}}\Big)^{3} \\
    & = 
    (\alpha_{1}^{*})^{6} 
    + 
    5(\alpha_{1}^{*})^{4}(\alpha_{2})^{1}
    +
    6(\alpha_{1}^{*})^{2}(\alpha_{2})^{2}
    +
    (\alpha_{2})^{3}
\end{aligned}
\end{equation}

This can also be generalized to larger $k$ values. Basically, comparing to vanilla RNN, DCRNN have more terms in its backpropagation and will better model higher order dynamical equations.

\end{document}